\documentclass[11pt, a4paper, logo, onecolumn, nonumbering]{SAIS}

\pdftrailerid{redacted}


\usepackage{kantlipsum, lipsum}
\usepackage{dsfont}
\usepackage{multirow}
\usepackage{array}
\usepackage{amssymb} %
\usepackage{pifont} %
\usepackage{makecell}
\usepackage{algorithm}
\usepackage{algpseudocode}
\usepackage{amsmath}
\usepackage{multicol}
\usepackage{float}
\usepackage{siunitx}
\usepackage{tabularx}
\usepackage{booktabs}

\usepackage{xcolor}
\definecolor{GoogleBlue}{HTML}{1A73E8}
\definecolor{GoogleBlueLight}{HTML}{e8f0fe}
\definecolor{GoogleBlueDark}{HTML}{174ea6}
\definecolor{GoogleRed}{HTML}{D93025}
\definecolor{GoogleGray}{HTML}{767676}
\definecolor{figtextcolor}{HTML}{333333}
\definecolor{perceptioncol}{HTML}{1A73E8}
\definecolor{modelingcol}{HTML}{925BCF}
\definecolor{manipulationcol}{HTML}{C440A7}
\definecolor{reasoningcol}{HTML}{DA2F77}

\usepackage{subcaption}
\usepackage{opensans}
\DeclareCaptionFont{subfigcapsize}{\fontsize{8pt}{8pt}\selectfont}  %
\usepackage{xurl}
\usepackage[colorlinks=true, allcolors=GoogleBlue]{hyperref}
\usepackage[capitalise]{cleveref}
\crefname{section}{Sec.}{Secs.}  %
\crefname{subsection}{Sec.}{Secs.}  %
\crefname{subsubsection}{Sec.}{Secs.}  %
\crefname{appendix}{App.}{Apps.}  %

\usepackage{graphicx}
\graphicspath{{figures/}}

\usepackage[breakable]{tcolorbox}

\newcounter{takeaway}
\usepackage{tikz}
\usetikzlibrary{calc}
\newlength{\tikzwidth}
\newlength{\boxpad}
\newlength{\boxgap}
\newlength{\boxtextheight}
\title{\centering Omni-Video 2: Scaling MLLM-Conditioned Diffusion for Unified Video Generation and Editing}
\usepackage[numbers, sort&compress]{natbib}
\usepackage{graphicx}
\correspondingauthor{my-email@google.com}

\reportnumber{} %

\author[2]{Hao~Yang}
\author[1,2]{Zhiyu~Tan\textsuperscript{$\dagger$}}
\author[2]{Jia~Gong}
\author[2]{Luozheng~Qin}
\author[1,2]{Hesen~Chen}
\author[2]{Xiaomeng~Yang}
\author[2]{Yuqing~Sun}
\author[2]{Yuetan~Lin}
\author[1,2]{Mengping~Yang\textsuperscript{*}}
\author[1,2]{Hao~Li\textsuperscript{*}}
\affil[1]{Fudan University}
\affil[2]{Shanghai Academy of Artificial Intelligence for Science \par $\dagger$Project Lead,  \textsuperscript{*}Corresponding Authors}

\newcommand{\websiteurl}{https://howellyoung-s.github.io/Omni-Video2-project/}

\newcommand{\github}{https://github.com/SAIS-FUXI/Omni-Video}

\newcommand{\HFmodel}{https://huggingface.co/Fudan-FUXI/OmniVideo2-A14B}

\begin{abstract}
Recent progress in unified modeling has shown great promise in handling diverse tasks including understanding and generating multimodal content, yet a crucial gap remains in applying this paradigm effectively to the video domain.
Specifically, the computational demands and temporal consistency requirements have made scaling video models notoriously difficult, while the lack of deep semantic grounding has also largely hindered the ability to follow precise editing instructions.
To bridge this gap, we present Omni-Video 2, a scalable and computationally efficient model that connects pretrained multimodal large-language models (MLLMs) with video diffusion models for unified video generation and editing.
Our key idea is to exploit the understanding and reasoning capabilities of MLLMs to produce explicit target captions to interpret user instructions.
In this way, the rich contextual representations from the understanding model are directly used to guide the generative process, thereby improving performance on complex and compositional editing.
Moreover, a lightweight adapter is developed to inject multimodal conditional tokens into pretrained text-to-video diffusion models, allowing maximum reuse of their powerful generative priors in a parameter-efficient manner.
Benefiting from these designs, we scale up Omni-Video 2 to a 14B video diffusion model on meticulously curated training data with quality, supporting high quality text-to-video generation and various video editing tasks such as object removal, addition, background change, complex motion editing, \emph{etc.}
We evaluate the performance of Omni-Video 2 on the FiVE benchmark for fine-grained video editing and the VBench benchmark for text-to-video generation.
The results demonstrate its superior ability to follow complex compositional instructions in video editing, while also achieving competitive or superior quality in video generation tasks.
We open-source our code and models, including a smaller 1.3B and a higher-performing 7B unified video generator, to facilitate further research and development in unified video modeling.

Project Page: \url{\websiteurl}

GitHub Code: \url{\github}

HuggingFace Model: \url{\HFmodel}

\end{abstract}

\begin{document}

\newpage
\maketitle

\section{Introduction}
Unified modeling~\cite{tong2024metamorph,ma2025janusflow,deng2025bagel,luo2025univid,chen2025blip3} has recently emerged as a powerful paradigm for building general-purpose systems capable of understanding and generating multimodal content, including text, images, video, and their combinations.
In the realm of unified video modeling, a growing body of work explores joint video generation and editing by coupling multimodal large language models (MLLMs) models with diffusion-based generative backbones~\cite{wei2025univideo}, while other efforts aim to unify heterogeneous video editing tasks via in-context conditioning~\cite{ye2025unic,ju2025editverse}.
Despite this progress, building a practical unified video model remains highly nontrivial:
Video modeling incurs substantial computational overhead and imposes strict temporal consistency requirements, rendering large-scale training and inference particularly challenging. 
More importantly, training high-quality video generation models from scratch is prohibitively expensive, typically requiring massive datasets, long training schedules, and specialized infrastructure.
As a result, most practical unified video systems must rely on strong pretrained text-to-video diffusion models as their generative foundation, making effective reuse and preservation of such priors a central concern.

Building upon a pretrained T2V diffusion model is appealing due to its robustness and high-quality synthesis capabilities. 
However, unified training inevitably introduces new conditional signals such as source reference videos, multimodal interleaved conditions, and editing-specific instructions, which were absent during pretraining.
Without an explicit anchoring mechanism, these new inputs can interfere with the pretrained conditioning interface, causing distribution drift and catastrophic degradation on the original T2V task.
Meanwhile, instruction-driven editing further raises the bar: beyond semantic alignment with editing prompts, successful edits must preserve identity, background structure, and temporal coherence, while consistently applying the desired change across frames.

\begin{figure}[t]
    \centering
    \includegraphics[width=\linewidth]{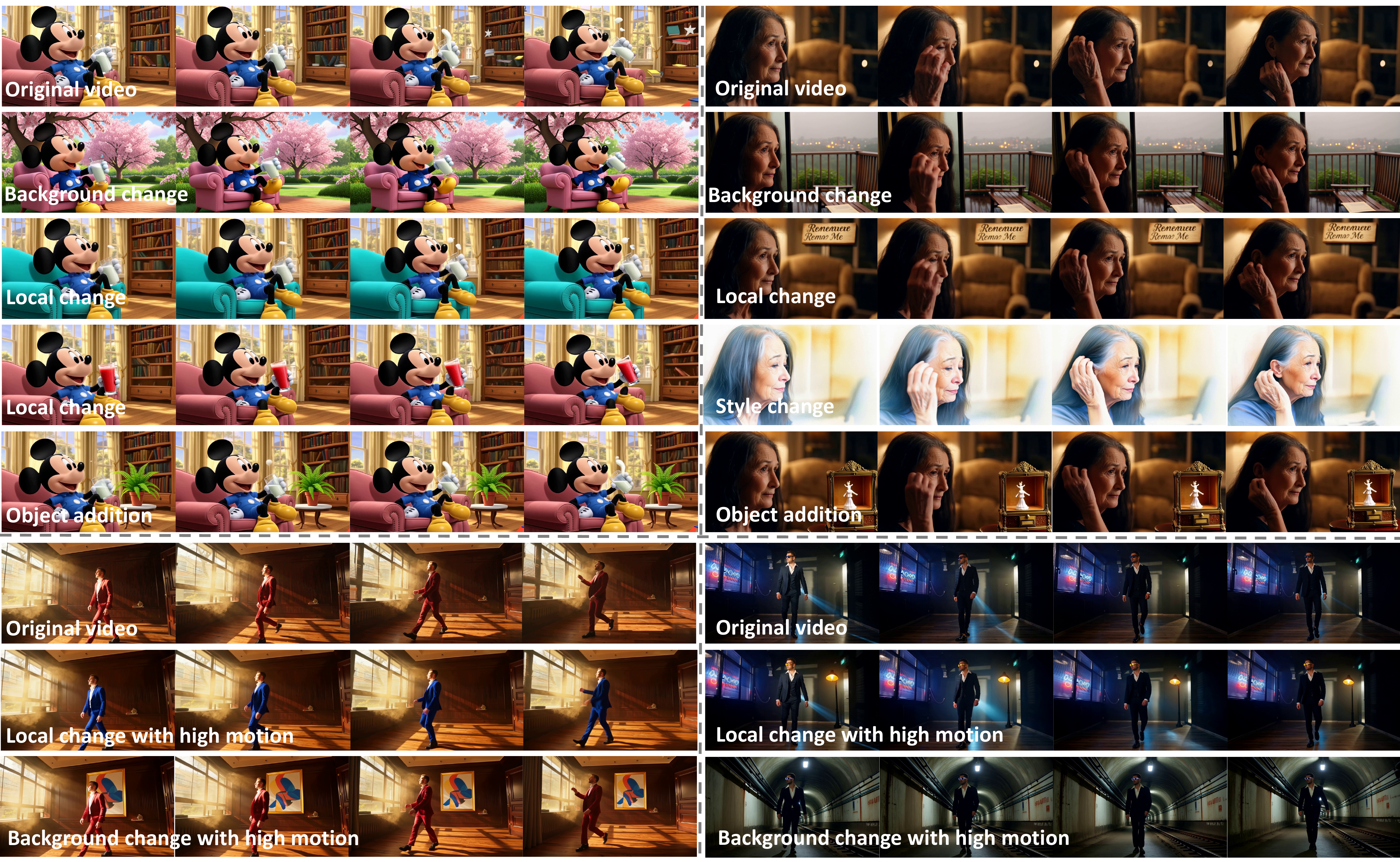}
    \caption{
    \textbf{We present Omni-Video 2, a unified video model for video generation and editing}.
    Our model supports various video editing tasks including local object changes (\emph{e.g., removal, addition}) and global changes (\emph{e.g., style, background}), producing edits that are both faithful to the user's prompt and temporally coherent with the original video, even when tested on videos with high motions.
    }
    \label{fig:teature}
\end{figure}

To address these challenges, we propose Omni-Video 2,  a scalable and computationally efficient unified video generation and editing framework built with a pragmatic objective: \emph{maximize reuse of a pretrained T2V foundation while enabling unified capabilities under limited additional training}.
Our central insight is that MLLMs possess strong understanding and reasoning capabilities that can be leveraged to interpret user intent more explicitly. 
Instead of directly conditioning video diffusion models on raw user prompts, Omni-Video 2 employs an MLLM-based editing prompt reasoner to generate explicit \emph{target captions} that precisely describe the desired visual outcome. 
These captions serve as structured semantic guidance, enabling the generative model to better follow complex and compositional editing instructions.
Even in the presence of additional controls (source references or editing instructions), the target caption continues to guide the diffusion process through the same interface learned during T2V pretraining. 
That is, new conditions are learned as \emph{additive refinements} to generation, rather than disruptive replacements of the pretrained guidance signal, thereby mitigating distribution drift and negative transfer.

To efficiently integrate multimodal conditions into video diffusion models while preserving pretrained behavior, we further introduce a lightweight multimodal condition adapter that injects conditional tokens into pretrained text-to-video diffusion backbones. %
Rather than modifying the core diffusion architecture or retraining the model end-to-end, the adapter provides a parameter-efficient pathway to encode additional control signals, such as source video references and editing-specific constraints.
This design minimizes interference with the original T2V conditioning mechanism, allowing the diffusion model to retain its pretrained generative prior while flexibly incorporating new modalities.
In this way, our model can fully exploit the powerful generative priors of existing diffusion models in a parameter-efficient manner, facilitating scalable training while maintaining high synthesis quality.

Building upon our earlier work on the Omni-Video v1~\cite{tan2025omni} model (MLLM‑7B paired with DiT‑1.3B) and the above architectural designs,  we scale Omni-Video 2 to a 14B video diffusion model trained on carefully curated, high-quality video data.
The resulting model supports high-quality text-to-video generation and a wide range of video editing tasks, including object removal and addition, background modification, and complex motion editing, \emph{etc.}.
Extensive experiments on the FiVE benchmark for fine-grained video editing and the VBench benchmark for text-to-video generation demonstrate that Omni-Video 2 achieves superior instruction-following capability while maintaining competitive or state-of-the-art generation quality.
To facilitate further research in unified video modeling, we open-source our code and models, including a compact 1.3B version and a higher-performing 7B unified video generator.

\section{Scaling Omni-Video 2}
\label{sec:method}


\subsection{Overview}
\label{sec:method_overview}

\begin{figure}[t]
    \centering
    \includegraphics[width=\linewidth]{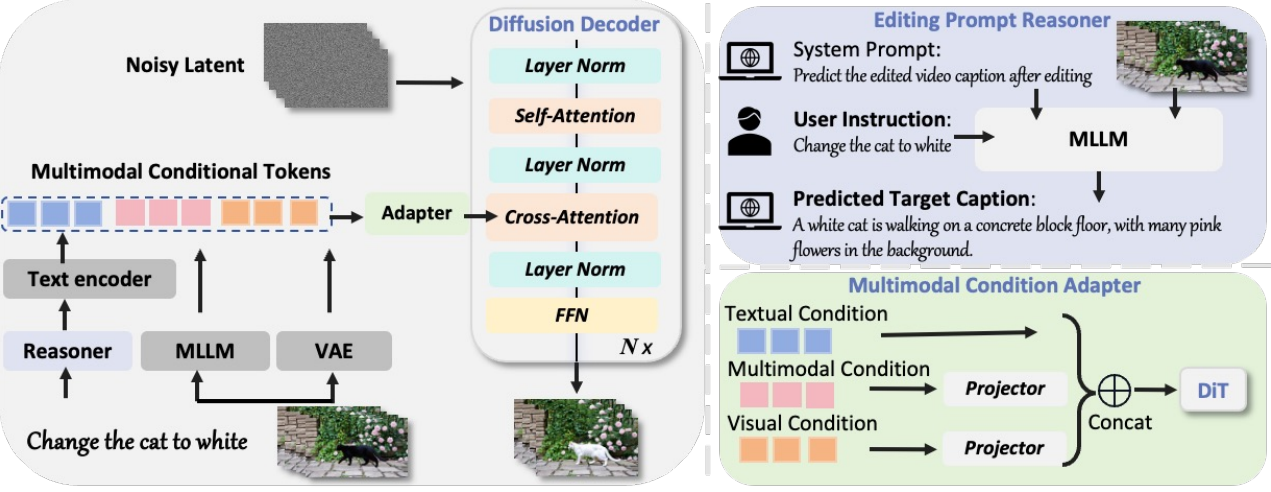}
    \caption{
    \textbf{Overall framework of Omni-Video 2 for unified video modeling.}
    An MLLM-based editing prompt reasoner first interprets the user instructions in the context of the source video to produce a precise target caption.
    A lightweight adapter then injects the multimodal conditional guidance into a powerful, pre-trained T2V diffusion model to perform editing. 
    Such design efficiently combines the MLLM's advanced reasoning with the T2V model's strong generative priors, enabling complex edits without costly full-model retraining.
    }
    \label{fig:omni_video2_framework}
\end{figure}

Omni-Video 2 is designed to support a broad spectrum of tasks, including text-to-image, text-to-video, and instruction-driven image and video editing, within a single unified architecture.
At a high level, Omni-Video 2 integrates an MLLM-based understanding branch with a diffusion-based generation branch into a unified modeling framework.
The understanding branch is responsible for multimodal perception, instruction interpretation, and high-level reasoning, while the diffusion decoder serves as the generative backbone that synthesizes visual content.
Rather than treating the MLLM and diffusion model as two loosely coupled components, Omni-Video 2 unifies them through explicit semantic interfaces. 
Concretely, as shown in Fig.~\ref{fig:omni_video2_framework}, our method is built upon two key components:
1) an Editing Prompt Reasoner, which leverages the understanding and reasoning capability of MLLMs to explicitly interpret user intent and produce structured semantic guidance; and
2) a Condition Adapter, which injects multimodal conditional signals into a pretrained diffusion backbone in a parameter-efficient and non-disruptive manner.
Together, these components enable Omni-Video‑2 to expand unified capabilities while maximizing reuse and preservation of a strong pretrained text-to-video (T2V) generative priors.

Formally, given an (optional) source video $x^{src}$ and an instruction $p^{edit}$, Omni-Video 2 first runs an MLLM to predict \emph{target video captions} $\hat{p}^{tgt}$ that explicitly specifies the desired output, and extracts \emph{cross-modal interaction features} that capture fused semantics from joint text-vision reasoning. 
These signals, together with a source-video VAE reference, form a  \emph{multimodal conditional tokens}, are then used to condition a pretrained diffusion-based video generator via a lightweight condition adapter.
Concretely, Omni-Video‑2 conditions the diffusion model on four sources:
\begin{enumerate}
    \item \textbf{MLLM target captions} $C^{tgt}$: a T5 embedding of the predicted caption $\hat{p}^{tgt}$, serving as an explicit constraint to maximize reuse of the pretrained text-to-video priors.
    \item \textbf{Editing instructions} $C^{edit}$: a T5 embedding of the original user instruction $p^{edit}$, retained to provide complementary instruction-level information.
    \item \textbf{MLLM cross-modal interaction features} $C^{\mathrm{MLLM}}$: a sequence of cross-modally fused tokens produced by the MLLM, capturing joint reasoning over instructions and visual inputs.
    \item \textbf{Source references} $C^{ref}$: VAE features of the source visual input (e.g., video latents), used as a persistent reference anchor to preserve identity, structure, and temporal coherence.
\end{enumerate}
By jointly conditioning on these signals and injecting them into the diffusion decoder with our developed adapter, Omni-Video‑2 enables precise and compositional instruction-following while treating newly introduced controls as additive refinements rather than replacements of the pretrained T2V prior.

\subsection{Editing Prompt Reasoner}
\label{sec:prompt_reasoner}
Omni-Video‑2 incorporates an \emph{Editing Prompt Reasoner} that leverages an MLLM to convert a user-provided edit instruction into an explicit, generation-friendly target caption. 
Given a source video $x^{src}$ and an edit instruction $p^{edit}$, the MLLM performs joint multimodal reasoning over the visual content and the instruction, and predicts a target video caption $\hat{p}^{tgt}$ that explicitly describes the intended output: $\hat{p}^{tgt} = \mathrm{MLLM}(x^{src}, p^{edit})$.
Such an explicit design addresses a key practical distributional shift in instruction-driven video editing:
Free-form edit prompts are often ambiguous, underspecified or implicitly dependent on the source content for context (e.g., referring to objects, attributes, or motions without explicitly naming them). 
While these free-form prompts are natural for human users, they form a suboptimal conditioning interface for pretrained text-to-video diffusion models, which are typically trained on explicit, self-contained and detailed captions. 
Consequently, directly conditioning on raw edit prompts increases uncertainty in the guidance signal and can lead to unstable or inconsistent edits.
By contrast, the Editing Prompt Reasoner uses the MLLM’s strong instruction understanding and reasoning capabilities to transform the raw instruction into a coherent and explicit caption that clearly specifies the desired visual outcome, while preserving relevant context from the source video.
More importantly, using the target caption as a primary guidance signal allows Omni-Video‑2 to retain the original caption-based conditioning pathway learned during text-to-video pretraining, maximizing reuse of the pretrained text-to-video prior and mitigating distribution drift and negative transfer.

\subsection{Multimodal Condition Adapter}
\label{sec:adapter}
To integrate heterogeneous conditional signals into the diffusion model in a unified and parameter-efficient manner, Omni-Video‑2 employs a Multimodal Condition Adapter that injects all conditions through a shared cross-attention interface (as shown in Fig.~\ref{fig:omni_video2_framework}). 
Rather than modifying the diffusion backbone or introducing task-specific heads, our design constructs a single unified conditioning sequence that serves as the keys and values for cross-attention in the diffusion transformer.
Specifically, we first obtain \emph{MLLM cross-modal interaction features} from the final-layer hidden states of MLLM: $H^{\mathrm{mllm}} \in \mathbb{R}^{L \times d_{\mathrm{mllm}}}$, where $L$ is the total number of multimodal condition tokens (including both text tokens and visual tokens of the source video), and $d_{\mathrm{mllm}}$ is the MLLM hidden dimension.
In practice, $L$ varies with input duration/resolution and prompt length; Omni-Video 2 therefore treats $H^{\mathrm{MLLM}}$ as a variable-length conditioning sequence. %
When computational or memory constraints require a fixed budget, we reduce the sequence to length $L_{mllm}$ using simple token selection strategies such as truncation (keeping the first/last $L_{mllm}$ tokens) or uniform sampling over visual tokens, and apply the same reduction consistently during training and inference.
The MLLM interaction features are aligned to the diffusion hidden space using a linear projector, $C^{mllm}=\mathrm{Projector}(H^{\mathrm{MLLM}})\in\mathbb{R}^{L_{mllm}\times d_{dit}}$.

In parallel, we encode the predicted target caption and the edit instruction with a T5 text encoder to obtain $C^{tgt}\in\mathbb{R}^{L_{tgt}\times d_{txt}}$ and $C^{edit}\in\mathbb{R}^{L_{edit}\times d_{txt}}$.
We apply learned linear projections to match the dimension of the diffusion decoder when necessary.
When a source input video is available, we additionally compute a VAE-based reference representation $C^{ref}$ from the source video latents.
All conditional tokens are then concatenated into a single conditioning sequence:
$C=[C^{mllm};\,C^{tgt};\,C^{edit};\,C^{ref}]\in\mathbb{R}^{L_C\times d_{dit}}$.
The unified sequence $C$ is used consistently across all DiT blocks as the source of keys and values in cross-attention, where latent tokens query $C$ via standard attention: $\mathrm{Attn}(Q,K,V)=\mathrm{softmax}(QK^\top/\sqrt{d})V$ with $K,V$ derived from $C$.
Unless otherwise stated, the same conditioning interface and adapter outputs are shared across all diffusion blocks, and we do not introduce block-specific injections.

\subsection{Multimodal condition Training}
\label{sec:method_multi_cond}
To improve robustness to missing or partial conditions and to prevent over-reliance on any single control signal, we adopt randomized condition dropout during training. 
Following the design principle of preserving the pretrained text-to-video prior, we treat the source VAE reference as a persistent structural anchor and apply dropout to all other conditioning sources. 
Specifically, we mask the MLLM interaction tokens, target condition, and edit instruction as
\begin{equation}
    \tilde{C}^{mllm} = m_{mllm}\cdot C^{mllm},\quad
    \tilde{C}^{tgt} = m_{tgt}\cdot C^{tgt},\quad
    \tilde{C}^{edit} = m_{edit}\cdot C^{edit},\quad
    \tilde{C}^{ref} = C^{ref},
\end{equation}
where $m\in\{0,1\}$ are independent Bernoulli masks with drop probabilities $p_{mllm}$, $p_{tgt}$, and $p_{edit}$. The final conditioning sequence is $\tilde{C}=[\tilde{C}^{mllm};\,\tilde{C}^{tgt};\,\tilde{C}^{edit};\,\tilde{C}^{ref}]$.
This strategy serves multiple purposes. 
First, it improves robustness at inference time, where some conditions (e.g., MLLM features or edit instructions) may be unavailable or intentionally omitted. 
Second, by occasionally dropping auxiliary conditions while retaining the pretrained caption pathway, the diffusion model is encouraged to maintain strong reliance on caption-based guidance, thereby reinforcing the pretrained text-to-video prior. 
Third, randomized dropout prevents the adapter from overfitting to high-dimensional MLLM features, stabilizing unified training across heterogeneous tasks.

Omni-Video‑2 unifies multiple generation and editing tasks by employing a shared diffusion decoder and a standardized four-condition interface, where task-specific inputs are treated as optional instantiations of the same conditioning structure. The instruction-based video editing formulation described above generalizes naturally to other tasks including text-to-image/video generation, instructional image/video editing.
Across all tasks, each sample is expressed using a standardized text template that populates 1) the edit instruction $p^{edit}$ and 2) the target caption $\hat{p}^{tgt}$, either predicted by the MLLM or provided by the dataset.
This standardization ensures a consistent conditioning format and reduces ambiguity in unified training.
We train on a mixture of datasets from these tasks and group samples into buckets by resolution and temporal length (number of frames) to efficiently support variable-size training. 
Regarding training objective, we train the generator using a flow-based objective in latent space.
Let $z$ denote the VAE latent of the target visual content and sample a continuous-time parameter $u\sim\mathcal{U}[0,1]$. 
We then construct an interpolated latent $z_u$ along a predefined path between noise and data (e.g., a linear interpolation between $z$ and $\epsilon\sim\mathcal{N}(0,I)$), and learn a conditional velocity (flow) field $v_\theta$ that transports samples along this path conditioned on $\tilde{C}$. 
Concretely, we minimize the loss between the predicted velocity and the corresponding target velocity:
\begin{equation}
    \mathcal{L}_{flow}=\mathbb{E}_{z,\epsilon,u}\left[\left\|v_\theta\!\left(z_u, u \mid \tilde{C}\right)-v^\star\!\left(z,\epsilon,u\right)\right\|_2^2\right],
\end{equation}
where $v^\star$ is determined by the chosen path parameterization.

\section{Data Curation}
\label{sec:data}

%
We construct a dedicated unified training corpus for Omni-Video‑2 that jointly supports generation and editing, with a particular emphasis on preserving a strong pretrained text-to-video prior while improving controllability for instruction-driven video editing. 
The final dataset contains over one million training instances spanning text-to-image, text-to-video, image editing, and video editing tasks.
Following the unified conditioning design described in Sec.~\ref{sec:method}, all training samples are organized using a standardized schema with optional fields.
This enables consistent conditioning across heterogeneous tasks and allows the diffusion decoder to be trained with a shared interface, thus stabilizing unified training and transferring generation capability to more complex video editing scenarios.

\begin{figure}[t]
    \centering
    \begin{minipage}{0.49\linewidth}
        \centering
        \includegraphics[width=\linewidth]{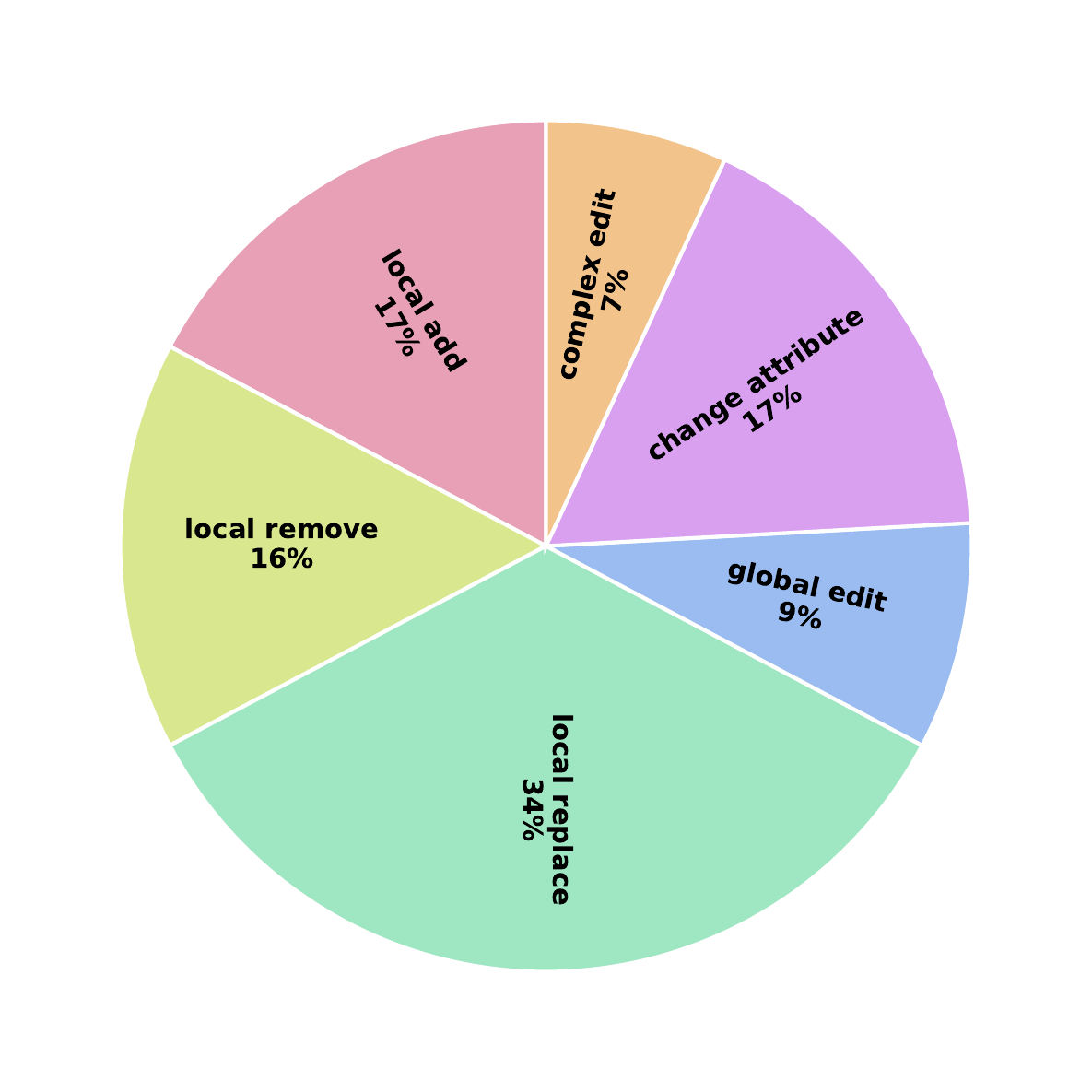}
        \captionof{figure}{
        \textbf{Video editing data instruction categories}.
        }
        \label{fig:edit_type_pie}
    \end{minipage}\hfill
    \begin{minipage}{0.49\linewidth}
        \centering
        \includegraphics[width=\linewidth]{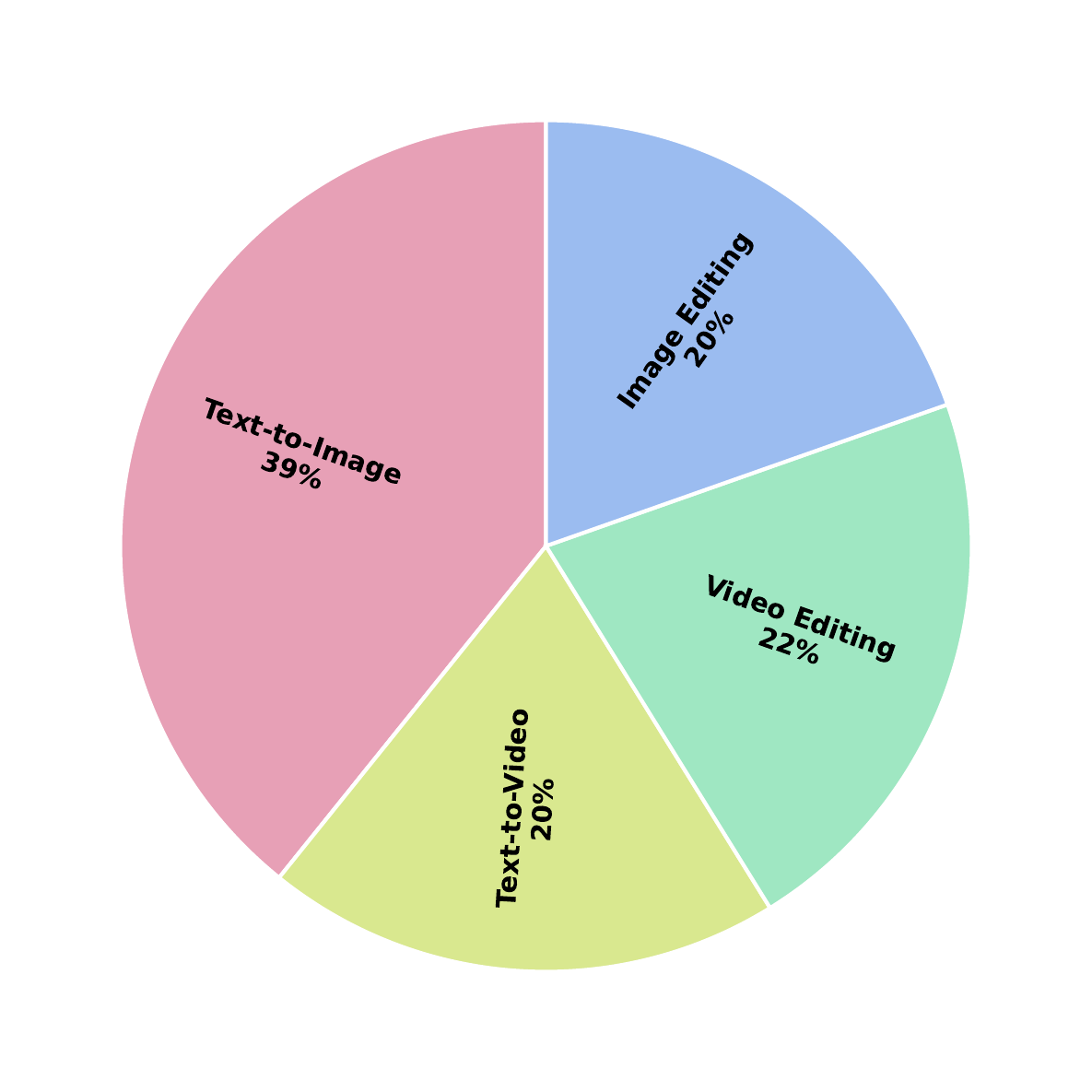}
        \captionof{figure}{
        \textbf{Composition of the final training dataset}.
        }
        \label{fig:data_task_pie} 
    \end{minipage}
\end{figure}

\subsection{Data Sources}
\label{sec:data_sources}
Our training data is collected from two complementary sources: real-world data and synthetic data, which play distinct but synergistic roles in the unified training pipeline. Real-world data provides diverse visual content and natural language descriptions drawn from in-the-wild distributions. This data is essential for learning rich appearance, motion, and scene dynamics, and for maintaining strong text-to-video generation quality. In particular, large-scale captioned videos serve as the backbone for preserving the pretrained text-to-video prior, ensuring that Omni-Video‑2 retains high-fidelity generation performance under caption-based guidance.

Synthetic data is primarily introduced to strengthen instruction-driven editing supervision, where naturally occurring data is often sparse, noisy, or ambiguously annotated. By synthetically constructing paired source–target examples with controlled edit operations, we obtain precise alignment between the edit instruction, the target caption, and the visual transformation. This is especially important for video editing, where fine-grained, temporally consistent edits (e.g., localized object changes or attribute modifications over time) are underrepresented in real-world datasets. Synthetic data thus complements real-world data by improving coverage of complex edit types and providing high-quality supervision for learning precise, localized video edits.

To strengthen instruction-driven editing, we curate and/or synthesize a diverse set of editing instructions and organize them into six categories: \textbf{local add}, \textbf{local remove}, \textbf{local replace}, \textbf{global edit}, \textbf{change attribute}, and \textbf{complex edit}. 
Local edits operate on confined regions or objects (adding, removing, or replacing entities) while preserving the remaining content; global edits apply scene-level transformations; attribute edits modify properties such as color, material, style, or motion; and complex edits combine multiple constraints (e.g., multi-object changes, multi-step instructions, or temporally persistent edits). Fig.~\ref{fig:edit_type_pie} illustrates the distribution of these editing categories.

\subsection{Unified Sample Format}
\label{sec:data_format}
To align with the four-condition interface used by Omni-Video‑2, we represent all tasks using a unified instance format with optional fields. Each training example contains a subset of the following components:
(i) a task type indicator (e.g., T2I/T2V/I2I/video-editing),
(ii) an optional source visual input (image or video) $x^{src}$,
(iii) an edit instruction $p^{edit}$ (which may simplify to a generation prompt for non-editing tasks),
(iv) a target caption $p^{tgt}$ (either provided or predicted by the Prompt Reasoner in Sec.~\ref{sec:prompt_reasoner}),
and (v) a target visual output (image or video).
Such unified representation directly mirrors the conditioning structure of the diffusion decoder. In particular, the explicit inclusion of a target caption—regardless of task type—ensures that all training instances are anchored to a caption-style semantic description. 
This design reinforces the caption-based guidance pathway during training and enables seamless transfer between generation and editing tasks, especially for video synthesis.

\subsection{Data Cleaning}
\label{sec:data_cleaning}
To ensure reliable supervision under unified multi-task training, we apply a multi-stage data cleaning pipeline to both real-world and synthetic data. 
The pipeline progressively removes corrupted samples, low-quality visual content, and inconsistent text–visual pairs, with additional verification steps tailored to editing data.

\textbf{Stage 1: Integrity and basic validity}.
We remove corrupted or undecodable files, samples with insufficient frame counts, and instances outside supported resolution or duration ranges. Near-duplicate detection is applied at the image and video level to reduce redundancy and mitigate leakage across dataset splits.

\textbf{Stage 2: Visual quality filtering}.
After the basic validity stage, we filter samples exhibiting severe compression artifacts, heavy overlays (e.g., prominent watermarks or subtitles), or degenerate motion patterns, such as near-static videos dominating the sequence. These checks rely on scalable heuristics based on decoding statistics and frame-level measurements, ensuring efficiency at scale.

\textbf{Stage 3: Text–visual consistency}.
In this stage, we remove samples with weak alignment between textual descriptions and visual content. 
For captioned data, we compare the provided caption with an automatically generated description of the visual input and discard mismatched pairs. 
For instruction-based editing, we additionally verify that the edit instruction and target caption are mutually consistent, avoiding contradictory supervision signals.

\textbf{Stage 4: Editing-specific verification}.
For paired editing instances, we apply extra validation to ensure that the intended modification is reflected in the target while non-edited content remains as consistent as possible. For local edits, we check that visual differences are spatially or temporally concentrated rather than globally disruptive. Difficult edit categories, such as multi-attribute or temporally persistent edits, are subject to additional spot checks to ensure supervision fidelity.

After cleaning, the final dataset contains over one million instances spanning four major task families: text-to-image, text-to-video, image editing, and video editing. 
Figure~\ref{fig:data_task_pie} summarizes the overall task composition.
To efficiently support variable-resolution and variable-length video training, we further group samples by spatial resolution and temporal duration (number of frames). 
These buckets are shared across all tasks, allowing unified batching and efficient utilization of training resources.
\section{Training}
\label{sec:Training}
%
We train OmniVideo‑2 with the explicit goal of maximally preserving the pretrained text-to-video prior while extending the model to support unified generation and instruction-driven editing.
A key design principle of OmniVideo‑2 is that all new capabilities are introduced through conditioning and lightweight adapter, rather than by altering the core diffusion architecture. 
The DiT backbone is initialized from a pretrained text-to-video model and remains structurally unchanged.
New conditional tokens (MLLM interaction features, edit instructions, and target captions) are injected exclusively through the shared cross-attention interface via lightweight adapters.
The MLLM remains frozen throughout training and is used solely for prompt reasoning and target-caption generation, ensuring that its instruction-following and multimodal reasoning abilities are fully preserved.
As a result, the diffusion model continues to operate in a familiar caption-conditioned regime, with additional conditions acting as auxiliary controls.
Consequently, only the diffusion decoder and the adapter are trainable in our implementation.

Specifically, OmniVideo‑2 consists of two 14B-parameter diffusion transformers (DiTs), corresponding to a low-noise model and a high-noise model. Training is conducted on 1600 Alibaba PPU GPUs using DeepSpeed with ZeRO‑1, mixed-precision training, gradient checkpointing, and standard large-scale system optimizations.
At the 14B scale, we find that simply increasing the number of GPUs and scaling up the batch size does not lead to faster convergence. In practice, large batch sizes significantly increase the wall-clock time per optimization step, reducing the number of parameter updates achievable within a fixed training budget. 
In our initial setup, even with 1600 GPUs, the system completed only approximately 4{,}000 steps per day, which noticeably slowed convergence. Standard heuristics, such as linear learning-rate scaling with batch size, did not yield meaningful improvements.

To overcome this bottleneck, we adopt Ulysses-style sequence parallelism and disable gradient checkpointing. 
Unlike typical diffusion setups, where sequence parallelism is applied mainly to self-attention, OmniVideo‑2 employs long and heterogeneous conditioning sequences due to the concatenation of MLLM features, target captions, edit instructions, and optional reference tokens. 
As a result, cross-attention constitutes a significant fraction of the total computation.
We therefore apply sequence parallelism to both self-attention and cross-attention layers in the DiT. 
With a sequence-parallel degree of 8, this design yields a $4$–$5\times$ speedup per training step, substantially increasing the effective update frequency and accelerating convergence. 
Importantly, this improvement allows us to maintain a relatively small batch size, which we find sufficient for stable training and beneficial for preserving the pretrained text-to-video dynamics.

\section{Evaluation}
\label{sec:qualitative}

\subsection{Evaluation Setup}
We evaluate Omni‑Video 2 on two fundamental and complementary tasks: video generation and video editing. 
Video generation measures the model’s ability to synthesize videos directly from textual prompts, while video editing evaluates its capability to perform text-guided modifications on existing videos.
For video generation, we benchmark Omni‑Video‑2 on VBench~\cite{huang2024vbench}, a widely adopted benchmark that evaluates video generation quality from multiple dimensions, including semantic alignment with text prompts, temporal coherence, motion naturalness, and overall visual fidelity.
Regarding video editing, we evaluate Omni‑Video 2 on FiVE‑Bench~\cite{li2025five}, a structured benchmark specifically designed for fine-grained video editing. 
FiVE‑Bench consists of high-quality source videos paired with editing instructions and target descriptions, covering six representative editing categories, including object replacement (rigid and non-rigid), color alteration, material modification, object addition, and object removal. 
This emphasis on localized, instruction-faithful, and temporally consistent edits directly aligns with the design goals of Omni‑Video‑2.

\begin{table}[t]
\centering
\caption{
\textbf{Quantitative comparison results of instructional video editing methods on the FiVE~\cite{li2025five} benchmark using FiVE-Acc metric}.
Omni-Video 2 significantly advances video editing performance compared with existing alternatives, suggesting a superior capability in accurately interpreting complex compositional instructions and translating them into temporally coherent video edits. 
}
\label{tab:basellines_five_acc}
\renewcommand{\arraystretch}{1.2}
\begin{tabular}{lccccc}
\toprule
\multicolumn{1}{l}{Model Name} &{FiVE-YN} &{FiVE-MC} &{FiVE-$\cup$} &{FiVE-$\cap$} & FiVE-Acc$\uparrow$ \\
\midrule
TokenFlow~\cite{geyer2024tokenflow}  &19.36 & 35.51 & 36.68 & 18.18 & 27.43 \\
DMT~\cite{yatim2024space}  & {34.78} & {62.06} & {62.98} &  {33.86} & {48.42} \\
VidToMe~\cite{li2024vidtome}   &20.03 & 33.50 & 36.20 & 17.34 & 26.77 \\
AnyV2V~\cite{ku2024anyv2v}  &30.62 & 45.42 & 48.96 & 27.09 & 38.02 \\
VideoGrain~\cite{yang2025videograin} &30.50 & 43.97 & 44.30 & 30.17 & 37.23 \\
Pyramid-Edit~\cite{li2025five}  &33.67 & {54.01} & {56.36} & 31.31 & 43.84 \\
Wan-Edit~\cite{li2025five}      & {41.41} & 52.53 & 55.72 & {38.22} & {46.97} \\
UniVideo~\cite{wei2025univideo}      & {56.50} & 68.55 & 69.95 & {55.10} & {62.53} \\
\midrule
\textbf{Omni-Video 2 (Ours)}   & \textbf{63.77} & \textbf{83.30} & \textbf{85.99} & \textbf{61.08} & \textbf{73.53}\\
\bottomrule
\end{tabular}
\end{table}

\subsection{Video Editing Performance}
\begin{figure}
    \centering
    \includegraphics[width=\linewidth]{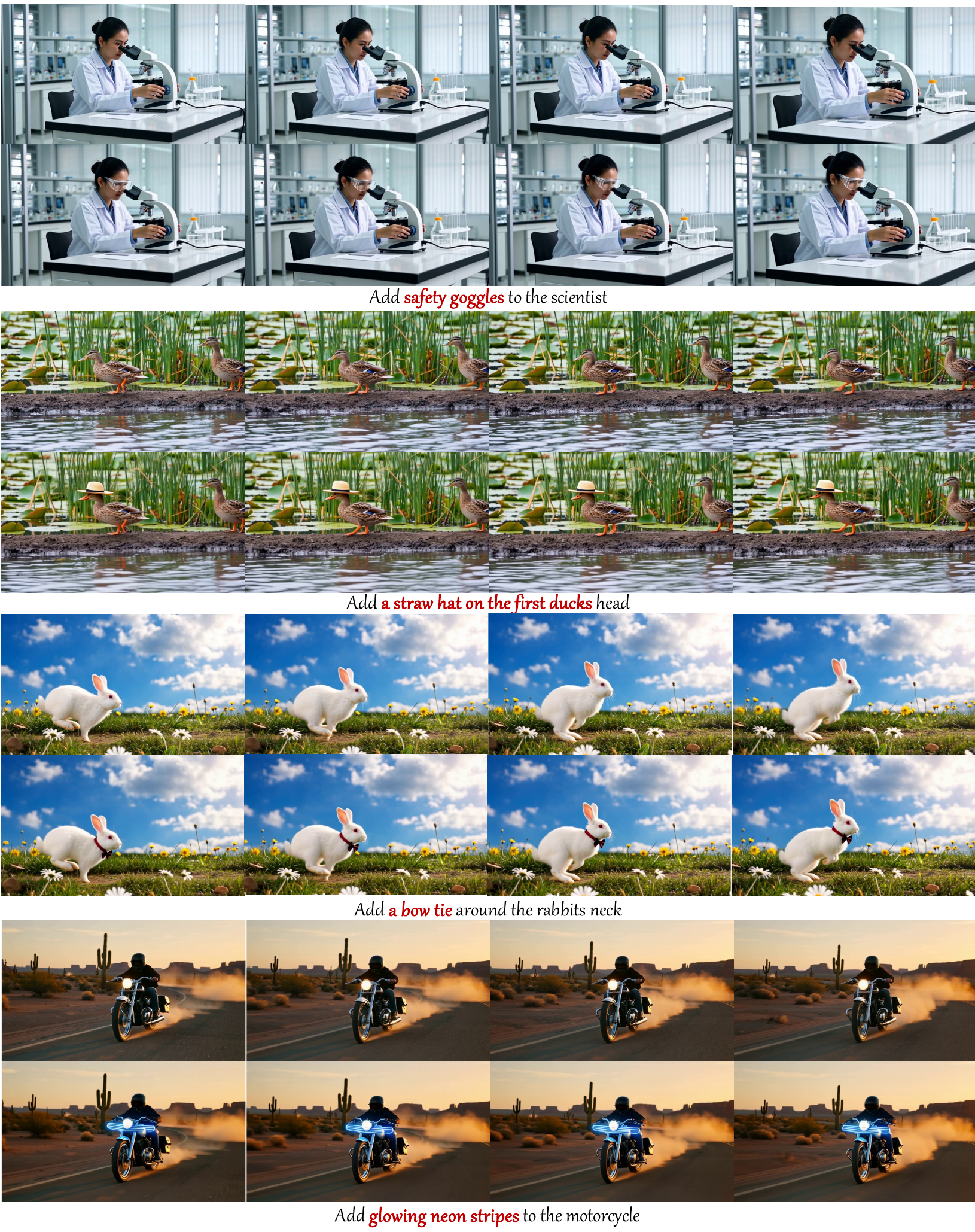}
    \caption{
    \textbf{Qualitative editing results on adding local object}.
    Omni-Video 2 accurately adds new objects based on the editing instructions while preserving the temporal consistency of the original video. The generated objects are realistic and well-integrated into the scene.
    }
    \label{fig:addition}
\end{figure}

\begin{figure}
    \centering
    \includegraphics[width=\linewidth]{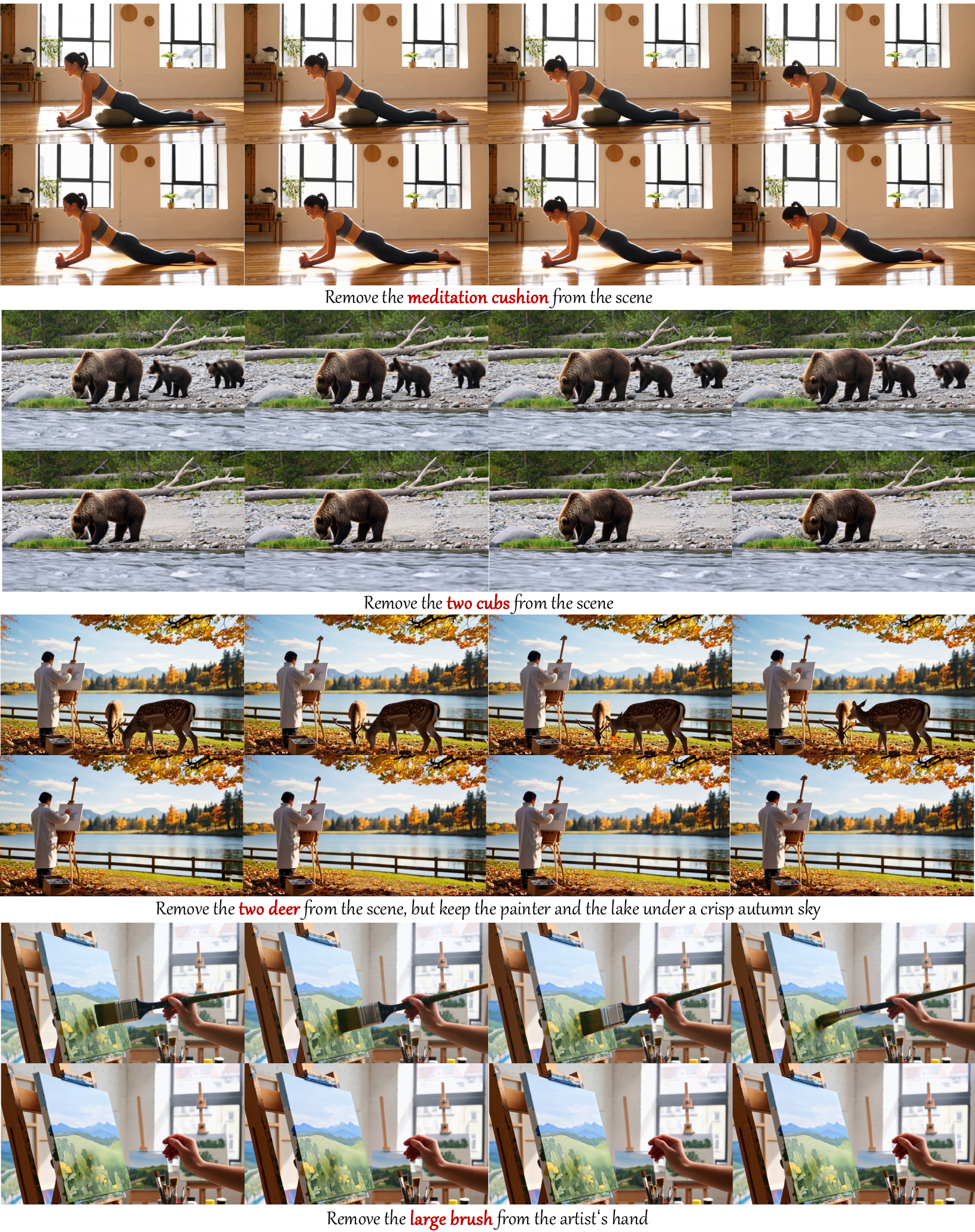}
    \caption{
    \textbf{Qualitative editing results on removing local object}.
    Omni-Video 2 successfully remove corresponding objects based on the editing instructions while preserving the temporal consistency of the original video.}
    \label{fig:remove}
\end{figure}

\begin{figure}
    \centering
    \includegraphics[width=\linewidth]{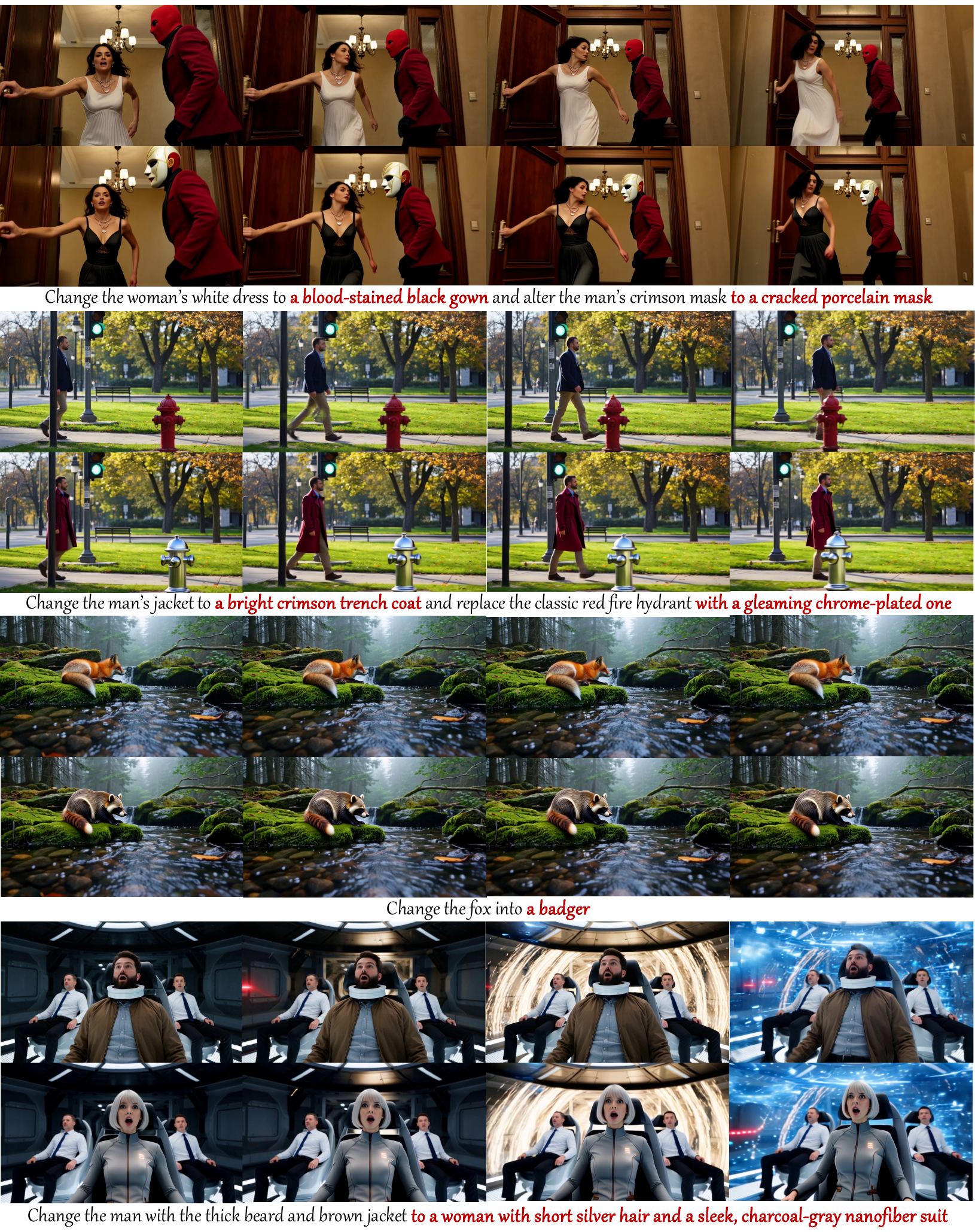}
    \caption{
    \textbf{Qualitative editing results on local changes}.
    Omni-Video 2 successfully change the attributes of corresponding local objects based on the editing instructions while preserving the temporal consistency of the original video.
    }
    \label{fig:local}
\end{figure}

\begin{figure}
    \centering
    \includegraphics[width=\linewidth]{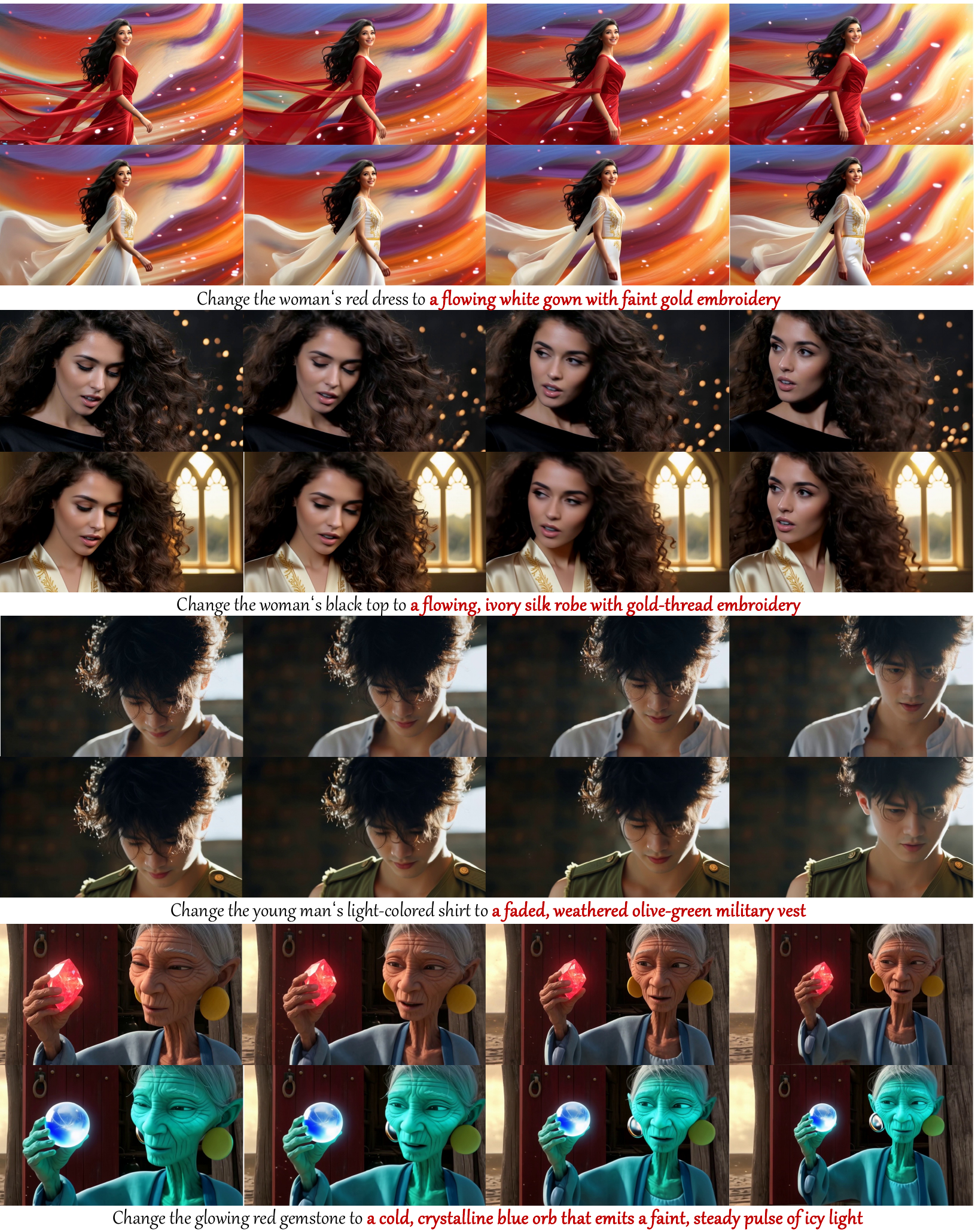}
    \caption{
    \textbf{Qualitative editing results on editing attributes of portraits}.
    Omni-Video 2 successfully change the attributes of portraits based on the editing instructions while preserving the temporal consistency of the original video.
    }
    \label{fig:potrait}
\end{figure}

\begin{figure}
    \centering
    \includegraphics[width=\linewidth]{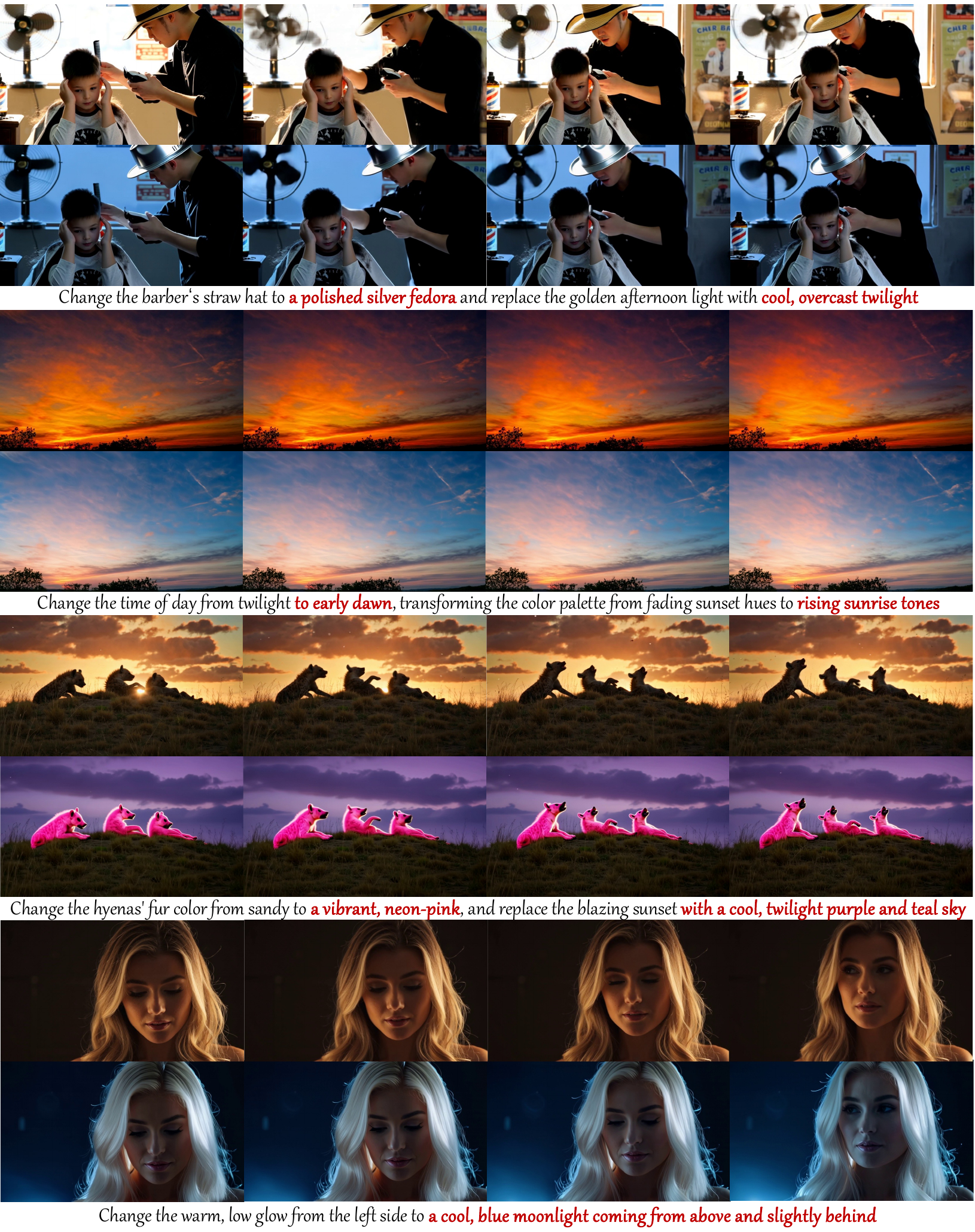}
    \caption{
    \textbf{Qualitative editing results on editing global attributes}.
    Omni-Video 2 successfully change the global style and background of input videos based on the editing instructions while preserving the temporal consistency of the original video.
    }
    \label{fig:global}
\end{figure}

\begin{figure}
    \centering
    \includegraphics[width=\linewidth]{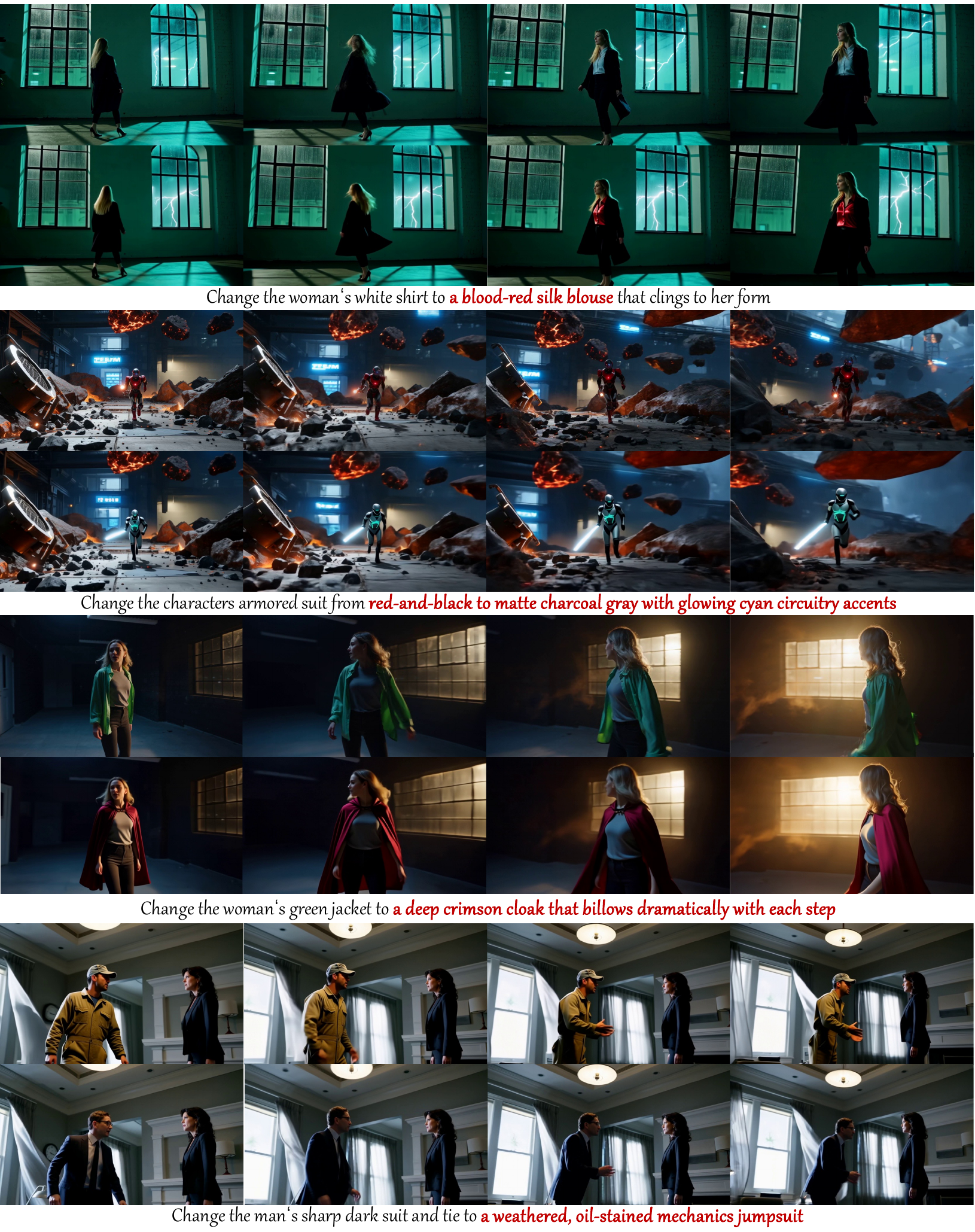}
    \caption{
    \textbf{Qualitative editing results on complex motion dynamics}.
    Omni-Video 2 demonstrates remarkable proficiency in editing videos characterized by complex dynamic object motion. The model not only executes the edit with high fidelity but also faithfully preserves the original motion patterns of the scene.
    }
    \label{fig:motion}
\end{figure}

\begin{figure}
    \centering
    \includegraphics[width=\linewidth]{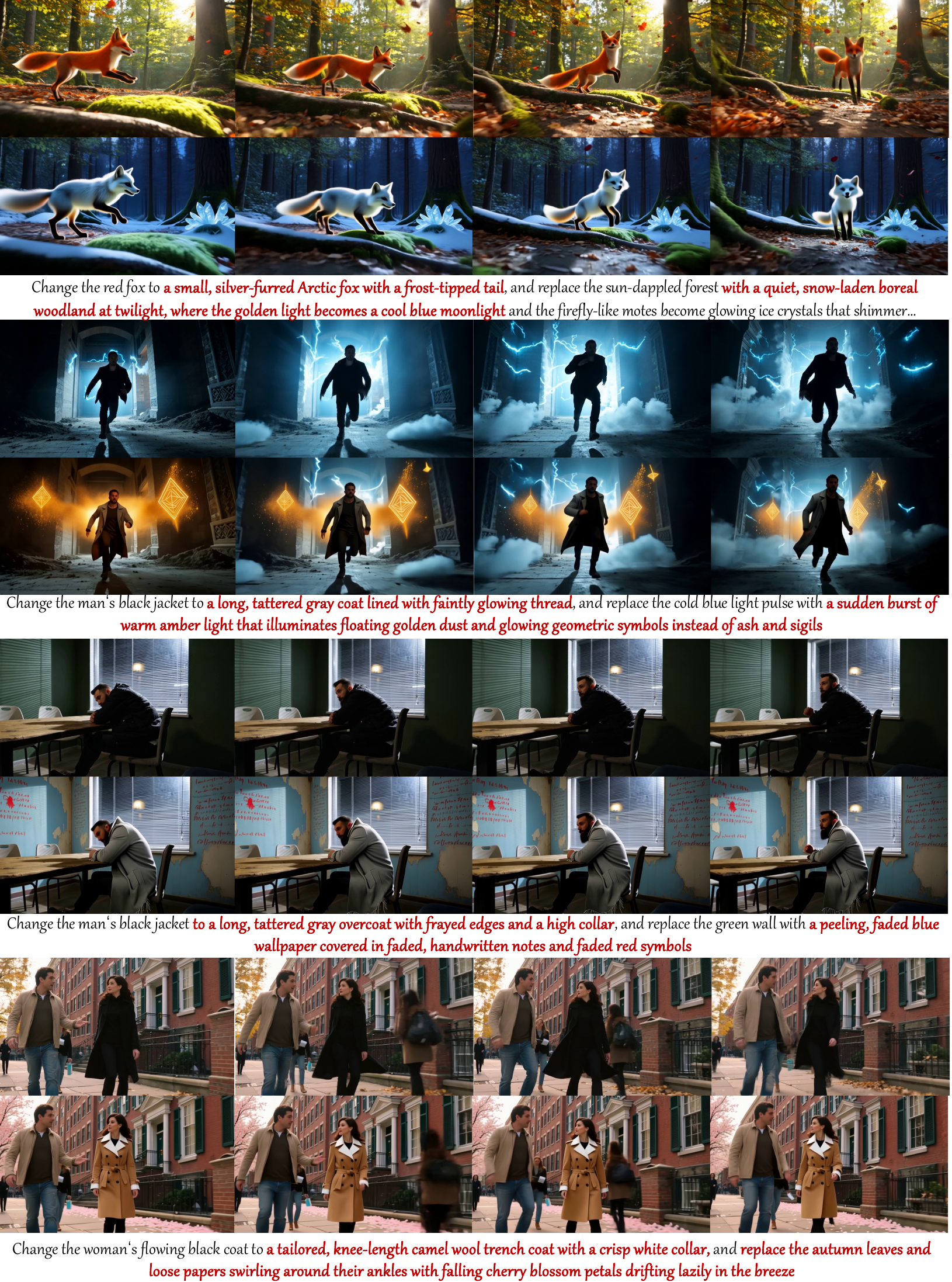}
    \caption{
    \textbf{Qualitative editing results on complex user instructions}.
    Omni-Video 2 excels at understanding and implementing complex, multi-part editing instructions. It successfully disentangles intricate concepts and relationships within the prompt to produce edits that are both accurate and temporally coherent.
    }
    \label{fig:complex}
\end{figure}

Tab.~\ref{tab:basellines_five_acc} shows the quantitative results on the video editing benchmark FiVE. 
It could be observed from the table that Omni‑Video‑2 significantly outperforms all existing baselines across all evaluation metrics, establishing a new state of the art on FiVE‑Bench.
Notably, Omni‑Video 2 achieves an overall FiVE‑Acc of 73.53, substantially surpassing the strongest prior method, UniVideo (62.53), by 11.0 absolute points.
This gap indicates a clear advantage in instruction-faithful and temporally consistent video editing, rather than incremental gains on isolated metrics.
More importantly, these gains are achieved without introducing task-specific architectures or fine-tuning. 
All editing capabilities emerge from the same diffusion backbone used for generation, confirming its ability to support high-quality video editing within a single unified framework.
Additionally, Fig.~\ref{fig:addition}--Fig.\ref{fig:complex} present qualitative editing results of Omni‑Video‑2 across a wide range of editing categories, including local object manipulation, attribute modification, global edits, motion-sensitive edits, and complex multi-part instructions.
These visualizations provide intuitive evidence that complements the quantitative improvements reported on FiVE‑Bench.

\begin{table*}[t]
\caption{
\textbf{Quantitative comparison results of text-to-video generation methods on the VBench~\cite{huang2024vbench} benchmark}.
Omni-Video 2 achieves strong performance, outperforming leading text-to-video generation methods across a majority of evaluation dimensions.
These results show that our framework for enabling complex editing does not compromise the foundational model's core generative capabilities.
Quantitative results are quoted from the official Vbench Leaderboard.
}
\label{tab:vbench_forced}
\resizebox{\textwidth}{!}{%
\begin{tabular}{lcccccccccc}
\toprule
\multicolumn{1}{c}{}                                                    & \multicolumn{3}{c}{\textbf{Overall Scores}}        & \multicolumn{5}{c}{\textbf{Technical Quality}}                                           & \multicolumn{2}{c}{\textbf{Aesthetic Quality}}         \\ \cmidrule(lr){2-4} \cmidrule(lr){5-9} \cmidrule(lr){10-11}
\textbf{Model Name}                                                     & Total Score    & Quality          & Semantic       & Subject & Background      & Temporal & Motion  & Dynamic & Aesthetic    & \multicolumn{1}{c}{Imaging} \\ \midrule
EasyAnimateV5.1~\cite{xu2025easyanimate}          & {83.42} & 85.03            & 77.01          & 98.00            & 97.41            & 99.19            & 98.02          & 57.15          & \textbf{69.48}       & \multicolumn{1}{c}{\textbf{68.61}}  \\
MiniMax-01~\cite{minimax2024video}          & 83.41          & 84.85            & 77.65          & 97.51            & 97.05            & 99.10            & {99.22}          & 64.91          & 63.03                & \multicolumn{1}{c}{67.17}           \\
Kling 1.6~\cite{kuaishou2025kling}                & 83.40          & 85.00            & 76.99          & 97.40            & 96.84            & 98.64            & 99.13          & 62.22          & 64.81                & \multicolumn{1}{c}{69.70}           \\
Wan2.1-T2V-1.3B~\cite{wan2025wan}                 & 83.31          & {85.23}   & 75.65          & \textbf{97.56}            & \textbf{97.93}   & \textbf{99.55}            & 98.52          & 65.19          & 65.46                & \multicolumn{1}{c}{67.01}           \\
HunyuanVideo~\cite{kong2024hunyuanvideo}          & 83.24          & 85.09            & 75.82          & 97.37            & 97.76            & 99.44            & 98.99          & {70.83} & 60.36                & \multicolumn{1}{c}{67.56}           \\
Gen-3~\cite{runway2024gen3}                       & 82.32          & 84.11            & 75.17          & 97.10            & 96.62            & 98.61            & \textbf{99.23} & 60.14          & 63.34                & \multicolumn{1}{c}{66.82}           \\
Vchitect-2.0 (VEnhancer)~\cite{fan2025vchitect}   & 82.24          & 83.54            & 77.06          & 96.83            & 96.66            & 98.57            & 98.98          & 63.89          & 60.41                & \multicolumn{1}{c}{65.35}           \\
CogVideoX1.5-5B~\cite{yang2024cogvideox}          & 82.17          & 82.78            & {79.76} & 96.87            & 97.35            & 98.88            & 98.31          & 50.93          & 62.79                & \multicolumn{1}{c}{65.02}           \\ \midrule
\textbf{Omni-Video 2 (Ours)}                                                           & \textbf{84.69}          & \textbf{85.79}           &\textbf{80.28} & 96.64   & 96.54            & 99.03   & 97.56 & \textbf{85.09}          & 66.48                & 67.36           \\ 
\midrule
\multicolumn{1}{c}{}                                                               & \multicolumn{8}{c}{\textbf{Semantic Fidelity}}                                                                                                                 \\ \cmidrule(lr){2-10}
\textbf{Model Name}          & Object   & Multi-Obj & Action   & Color            & Spatial & Scene            & Appearance  & Temporal & Overall   &                               \\ \midrule
EasyAnimateV5.1~\cite{xu2025easyanimate}          & 89.57          & 66.85            & 95.60          & 77.86            & 76.11            & 54.31            & 23.06          & 24.61          & 26.47                   &                          \\
MiniMax-Video-01~\cite{minimax2024video}          & {87.83} & {76.04}   & 92.40          & 90.36            & 75.50            & 50.68            & 20.06          & \textbf{25.63}          & 27.10                     &                           \\
Kling 1.6~\cite{kuaishou2025kling}                & 93.34          & 63.99            & 96.20          & 81.26            & 79.08            & 55.57            & 20.75          & 24.51          & 26.04                &                                     \\
Wan2.1-T2V-1.3B~\cite{wan2025wan}                 & 88.81          & 74.83            & 94.00          & 89.20            & 73.04            & 41.96            & 21.81          & 23.13          & 25.50               &                                \\
HunyuanVideo~\cite{kong2024hunyuanvideo}          & 86.10          & 68.55            & {94.40} & \textbf{91.60}   & 68.68            & 53.88            & 19.80          & 23.89          & 26.44                &                                  \\
Gen-3~\cite{runway2024gen3}                       & 87.81          & 53.64            & 96.40          & 80.90            & 65.09            & 54.57            & {24.31}          & 24.71          & 26.69             &                            \\
Vchitect-2.0 (VEnhancer)~\cite{fan2025vchitect}   & 86.61          & {68.84}   & {97.20} & 87.04            & {57.55}   & \textbf{56.57}   & 23.73          & {25.01} & {27.57}                      &            \\
CogVideoX1.5-5B~\cite{yang2024cogvideox} & {87.47} & {69.65}   & {97.20} & {87.55}   & \textbf{80.25}   & {52.91}   & \textbf{24.89} & {25.19} & 27.30                                  &            \\ \midrule
\textbf{Omni-Video 2 (Ours)}                                                           & \textbf{95.85} & \textbf{78.69}            & \textbf{97.60}          & 80.51   & 78.55   & 56.12   & 22.49 & 24.93 & \textbf{27.61}  & \\
\bottomrule
\end{tabular}
}
\end{table*}

\subsection{Video Generation Performance}
We further evaluate Omni‑Video‑2 on video generation using VBench, comparing it with state-of-the-art open-source and proprietary models. Quantitative results are summarized in Tab.~\ref{tab:vbench_forced}.
Despite being extended to support instruction-driven editing, Omni‑Video 2 achieves competitive or superior performance on standard video generation metrics, including semantic alignment, temporal coherence, and overall visual quality. 
Notably, the model exhibits no degradation in generation quality compared to strong text-to-video baselines. This stability demonstrates that the introduction of additional conditioning signals does not compromise the pretrained generation prior.
This result directly validates one of our core design principles: all extensions are introduced via lightweight conditioning adapters and unified training, without altering the core diffusion architecture. 
As a result, Omni‑Video‑2 retains the strengths of the original text-to-video model while gaining new capabilities.

Taken together, the results on video generation and video editing demonstrate that Omni‑Video 2 successfully unifies two traditionally separate paradigms within a single model. 
Unlike prior approaches that specialize in either generation or editing, Omni‑Video‑2 achieves strong performance on both tasks without the architectural branching or task-specific optimizations that characterize such specialized models.

\newpage

\section*{Acknowledgements}
The authors appreciate Shanghai Academy of Al for Science (SAIS) for computational resources, the Wan team for opensourcing the Wan series foundational video generation model, and the Qwen team for opensourcing the Qwen series multimodal large language model. 
Moreover, this report is build from the Latex template of Google Deepmind.
\bibliographystyle{abbrvnat}
\bibliography{ref}

\end{document}